\documentclass[10pt,conference,a4paper]{IEEEtran}
\IEEEoverridecommandlockouts
\usepackage{epsfig}
\usepackage{graphicx}
\usepackage{amsmath}
\usepackage{amssymb}
\usepackage{multirow}
\usepackage{multicol}
\graphicspath{./Images}
\usepackage{cite}
\usepackage{amsmath,amssymb,amsfonts}
\usepackage{algorithmic}
\usepackage{graphicx}
\usepackage{textcomp}
\usepackage{xcolor}
\usepackage{comment}

\def\BibTeX{{\rm B\kern-.05em{\sc i\kern-.025em b}\kern-.08em
    T\kern-.1667em\lower.7ex\hbox{E}\kern-.125emX}}
\begin{document}

\title{Generalized Iris Presentation Attack Detection Algorithm under Cross-Database Settings}

\author{
\IEEEauthorblockN{Mehak Gupta$^{1}$, Vishal Singh$^{1}$, Akshay Agarwal$^{1,2}$, Mayank Vatsa$^{3}$, and Richa Singh$^{3}$}
\IEEEauthorblockA{$^{1}$IIIT-Delhi, India; $^{2}$Texas A\&M University, Kingsville, USA; $^{3}$IIT Jodhpur, India\\ $^{1}$\{mehak16163, vishal16277, akshaya\}@iiitd.ac.in; $^{3}$\{mvatsa, richa\}@iitj.ac.in
}
}

\maketitle

\begin{abstract}

Presentation attacks are posing major challenges to most of the biometric modalities. Iris recognition, which is considered as one of the most accurate biometric modality for person identification, has also been shown to be vulnerable to advanced presentation attacks such as 3D contact lenses and textured lens. While in the literature, several presentation attack detection (PAD) algorithms are presented; a significant limitation is the generalizability against an unseen database, unseen sensor, and different imaging environment. To address this challenge, we propose a generalized deep learning-based PAD network, MVANet, which utilizes multiple representation layers. It is inspired by the simplicity and success of hybrid algorithm or fusion of multiple detection networks. The computational complexity is an essential factor in training deep neural networks; therefore, to reduce the computational complexity while learning multiple feature representation layers, a fixed base model has been used. The performance of the proposed network is demonstrated on multiple databases such as IIITD-WVU MUIPA and IIITD-CLI databases under cross-database training-testing settings, to assess the generalizability of the proposed algorithm.

\end{abstract}


%
\IEEEpeerreviewmaketitle

\section{Introduction}

Iris is considered one of the most accurate biometric modality for person recognition. The false accept rate of iris matching is considered to be the lowest among all other popular modalities, such as face and fingerprint \cite{jain2007handbook}. 
After the successful implementation of iris recognition for controlled applications including smartphone unlocking, researchers have been developing the technology to be implemented for semi-controlled and uncontrolled applications such as automatic access through airports\footnote{https://tinyurl.com/t9b4qfx} and securing the online wallets\footnote{https://tinyurl.com/v8pz6dp}. The usage of iris recognition is also explored in postmortem images \cite{trokielewicz2018iris, trokielewicz2020post}. However, person identification systems based on biometric recognition, including iris recognition, are vulnerable to presentation attacks. 


The effectiveness of attacks on iris recognition systems were highlighted when attackers successfully demonstrated the vulnerability of iris recognition systems on one of the popular mobile system\footnote{https://tinyurl.com/wssberz}. The images acquired with presentation attacks can serve two purpose: (i) identity evasion and (ii) identity impersonation. In the first case, the attacker can successfully hide his/her identity using presentation attack instruments (PAIs). In the second case, an attacker can assume the identity of someone else through different kinds of presentation attack instruments. The popular PAIs on iris recognition systems include 3D contact lens, iris image printouts, prosthetic eyes, and cadaver eyes. Printout based PAIs are generally of low image quality, contain texture artifacts such as Moir{\'e} pattern and reflection, and lack 3D structure. Due to these limitations, printout based attacks are easy to detect. Fake iris images based on contact lenses are rich in texture, have a 3D structure, and can quickly move with the real irises. Therefore, detection of these attacked images needs to be addressed more enthusiastically as compared to other PAIs. Researchers have also shown that contact lens-based iris images can degrade the iris recognition accuracy significantly \cite{baker2010degradation,kohli2013revisiting,yadav2014unraveling}. In the literature, numerous presentation attack detection (PAD) algorithms are presented which are either based on the extraction of texture features or motion features or deep learning features. However, recent iris PAD competition \cite{yambay2017livdet} have demonstrated limited generalizability against unseen databases and sensors. For instance, the best performing algorithm is not able to detect at least 38\% of PAIs in such settings.

\begin{figure}[]
\begin{center}
   \includegraphics[width=1\linewidth]{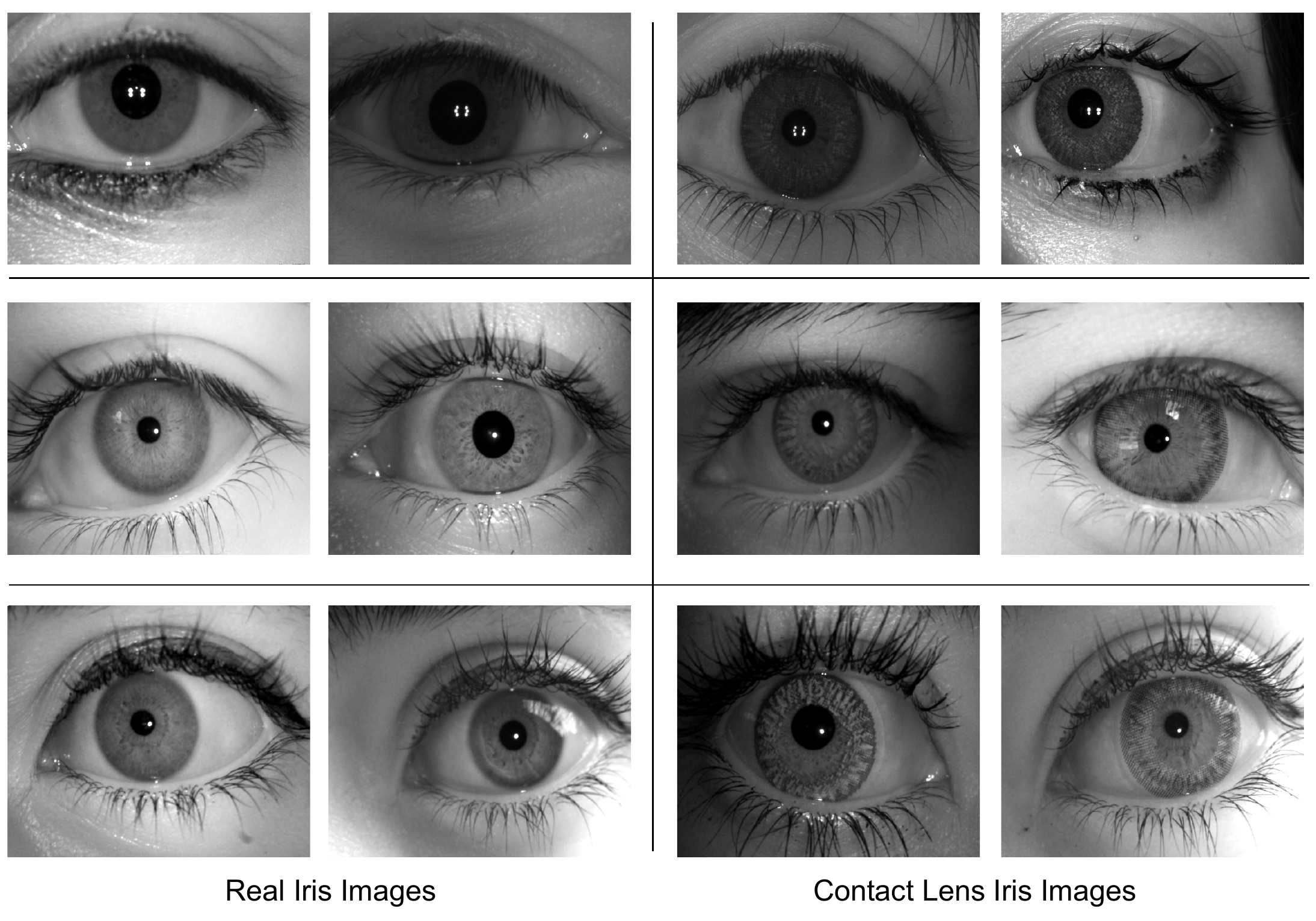}
\end{center}
\caption{Real and contact lens iris images from different databases. These examples showcase the variations due to illumination, contact lens type, and sensors used in the acquisition. Each row represents the images of one iris database.}
\label{fig:real-adv-samples}
\end{figure}

Motivated by the need of an efficient and robust algorithm, we present the proposed MVANet which utilizes the deep convolutional neural network architecture for presentation attack detection. In the traditional CNN models, a fully connected (FC) layer is bound with the last convolutional/pooling layer. Then several (optional) FC layers are connected sequentially with that FC layer. In the proposed algorithm, we have joined several FC layers parallelly with the last convolutional layer to learn multiple feature representations of an image. These layers are further combined with the softmax layer to train the model end-to-end through gradient accumulation and bifurcation. To evaluate the generalizability of the proposed algorithm, we have performed experiments using unseen training and testing scenarios. Three challenging presentation attack databases are used to conduct the experiments in three folds. In each fold, images from only one database are used for training, and images from the remaining two databases are used for testing. The performance of MVANet is also compared with three state-of-the-art (SOTA) CNN models including DenseNet \cite{DBLP:journals/corr/HuangLW16a}, ResNet18 \cite{DBLP:journals/corr/HeZRS15}, and VGG16 \cite{DBLP:journals/corr/SimonyanZ14a}, which are fine-tuned for iris PAD.

\subsection{Related Work}

The majority of iris presentation attack detection algorithms utilize hand-crafted features to encode the textural or morphological features of an iris image. One of the first iris PAD algorithms was proposed by Daugman utilizing the Fourier information from images. After that, several image-features based PAD algorithms are developed which yield high accuracy in constrained experimental settings. Gupta et al. \cite{gupta2014iris} have used several texture-based features such as local binary pattern (LBP), histogram of oriented gradients (HOG), and GIST. Raghavendra and Busch \cite{raghavendra2014presentation} have computed multiple features utilizing the multi-scale binarized statistical image features (BSIF) for presentation attack detection. Other popular images features used are DAISY \cite{tola2009daisy}, variants of LBP \cite{kohli2016detecting,nosaka2011feature}, scale-invariant descriptors (SID) \cite{doyle2015robust}. Several researchers, \cite{gragnaniello2015investigation,gragnaniello2015iris,hu2016iris}, have conducted studies using multiple image features and have shown interesting results. The quality and dimension of the features also depend on the region of the image used for its extraction. Some researchers have extracted the features from the full image while others have divided the image into multiple local patches and extracted the features. Gragnaniello et al. \cite{gragnaniello2016using} used both the iris region and the sclera region for the detection of contact lens-based presentation attacks. Bag of feature method is utilized on the features extracted from both the regions and linear support vector machine (SVM) classifier is trained for binary class classification. Furthermore, several researchers have also explored the motion features of iris and pupil regions for the identification of attacks \cite{czajka2015pupil,komogortsev2015attack,madhe2020design,rigas2015eye}. However, a significant limitation of the work is the degradation of pupillary light with time and with consumption of alcohol and drugs \cite{ssr}. Another limitation is that these methods can be circumvented using 3D attacks, which can move with real iris and pupil regions.

Utilizing the effectiveness of the convolutional neural network (CNN) in handling various image classification and detection tasks, researchers have started exploring it for iris PAD tasks as well. Menotti et al. \cite{menotti2015deep} have proposed the CNN model for various biometrics presentation attack detection including the face, fingerprint, and iris. Hoffman et al. \cite{hoffman2018convolutional,hoffman2019iris+} have extracted the features from the patches of the iris region and corresponding segmentation task. The information learned over multiple patches is fused for improved performance. Pala et al. \cite{pala2017iris} have used a five-layer CNN model and trained on three tuple information using the triplet loss. Chen and Ross \cite{chen2018multi} have used a similar concept while training the CNN model for the joint task of iris segmentation and PAD. He et al. \cite{he2016multi} trained the CNN models on 28 iris patches, and decision level fusion is performed for final classification. The significant limitation is the training complexity because of training the CNN individually over each patch. Choudhary et al. \cite{choudhary2019approach}, Singh et al. \cite{singh2018ghclnet}, and Yadav et al. \cite{yadav2019detecting} have used the ResNet and DenseNet models for contact lens detection. McGrath et al. \cite{mcgrath2018open} have developed an open-source iris presentation attack detection module utilizing publicly available machine learning feature extraction and classification algorithms.

\begin{figure*}[t]
\begin{center}
   \includegraphics[width=1.0\linewidth]{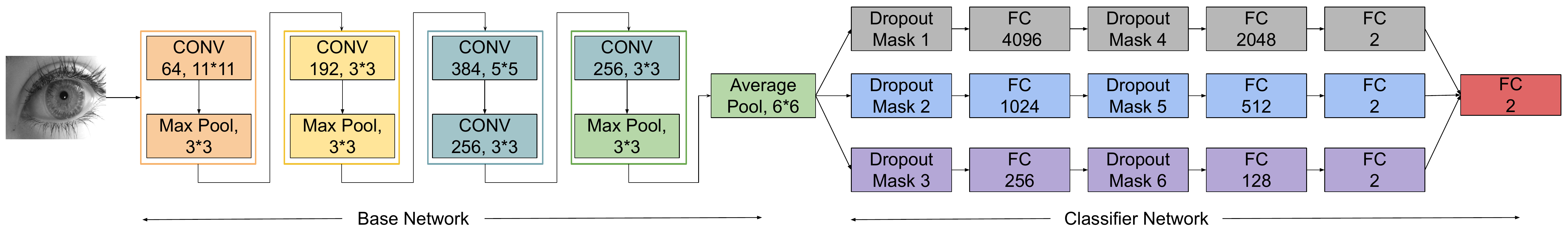}
\end{center}
\caption{Proposed MVANet for iris presentation attack detection in generalized settings.}
\label{fig:real-adv-samples}
\end{figure*}


Fang et al. \cite{fang2020robust} have used both 2D and 3D iris information for the detection of presentation attacks. The 2D information is learned using the ensemble of multiple textural features, and 3D shape features are learned using photometric stereo. Yadav et al. \cite{yadav2018fusion} and Choudhary et al. \cite{choudhary2020biometric} have combined the hand-crafted features computed over wavelet domain and CNN features for enhanced PAD. Czajka and Bowyer \cite{czajka2018presentation} and Chen and Zhang \cite{chen2018iris} have presented a comprehensive survey of most of the existing PAD algorithms. Due to the popularity of iris recognition and its sensitivity against presentation attacks several liveness detection competitions have been conducted. The first international competition was held in 2013 \cite{david2014livdet} and later two more competitions are conducted in 2015 \cite{david2017livdet} and in 2017 \cite{yambay2017livdet}. The recently presented survey papers and conducted competitions show that the effectiveness of the PAD algorithms has increased significantly in a constrained setting, including seen databases and seen sensors scenarios. The new algorithms need to be adapted to an unseen database, unseen sensor conditions, and demand that the future PAD algorithms must be tested against such generalized conditions \cite{yambay2019review}.

\section{Proposed Algorithm for Iris Presentation Attack Detection}

In this section, we describe the proposed deep learning based algorithm which can classify iris images as \textit{real} or \textit{attack}. It can be observed from the literature \cite{agarwal2016fingerprint, czajka2018presentation,hoffman2019iris+,peng2020face,siddiqui2016face,yadav2018fusion,agarwal2019iris,agarwal2016face} that hybrid algorithms which utilize multiple features or classifiers are highly successful and generalizable as compared to single classifier or feature-based algorithm. However, a significant drawback of the existing hybrid algorithms is that they either learn the separate CNN classifiers, hand-crafted features, or a combination of both, which is time-consuming. Therefore, to overcome such limitations of learning separate descriptors or classifiers from different algorithms, we have used the same base network while learning different representations. 

\subsection{Proposed CNN Architecture: MVANet}

The traditional CNN architecture either utilizes single or multiple fully connected layers to learn the features for classification. However, these layers are connected sequentially, i.e., only one layer is connected to base non-linear convolutional or pooling feature maps. Let us suppose if the CNN model consists of two fully connected (FC) layers: the transition of the layers in traditional CNN is represented as follows: $x_1 = H_{0}(\cdot)$, where $x_1$ is the first FC layer and $H_0$ represents one of the following layers: convolution (conv), pooling, ReLU, or Batch Normalization. The second FC layer ($x_2$) is connected to the first FC layer as follows: $x_2 = F_{x_1}(\cdot)$, where $F(\cdot)$ is the mapping function from one layer to another. In the deeper architecture (with multiple FC layers), due to vanishing gradient, it might be possible that the conv feature maps are not updated as desired for the classification task. In this research, we apply the FC layers in a parallel fashion compared to the traditional sequential fashion to reduce the vanishing gradient impact and ensure a smooth flow of information. 
The parallel connection of FC layers also helps in learning multiple representations of an image. The multiple FC layer representations can be written as: $x_{1...N} = H_{0}(\cdot)$, where $1...N$ represents the number of FC layer branches. In the proposed network, the value of $N$ is set to $3$, and $H_0$ is the average pooling layer. The detailed architecture is described below.

\subsubsection{Base Network}
The base network has five convolutional layers, with the first layer having filters of size $11\times11$, whereas the remaining layers use filters of size $3\times3$. Each of these convolutional blocks comprise convolution followed by Rectified Linear Unit (ReLU) and then a Batch Normalization (BN). Three Max Pool layers of $3\times3$ kernel size are present in between these convolutional layers. An average pooling follows all these layers, with the patch size being $6\times6$. The small base network makes the architecture time-efficient.

\subsubsection{Classifier Network}
The second part of the architecture is a multi-branch classifier network. It is based on the concept of multi-sample dropouts\cite{inoue2019multisample}, where each branch has a different set of weights. The output of the average pooling layer from the base network is fed into the three different classification branches, each of which is preceded by a dropout layer. The mask for the dropout layer in each branch is different but with same dropout rate of $0.5$. Because of this dropout layer in each classifier, we now have multiple dropout samples for each image instead of just one as is the case in a traditional dropout. All three classification layers contain three fully connected (FC) layers. 
\begin{itemize}
    \item The first classifier branch has a fully connected layer with $2048$ nodes, followed by a dropout layer. This is further connected to an FC layer with $1024$ nodes and finally to another fully connected layer with $2$ nodes.
    \item The second classifier branch has a similar structure having fully connected layers with $1024$, $512$, and $2$ nodes.
    \item Similarly, the third classifier branch has fully connected layers with $256$, $128$, and $2$ nodes.
\end{itemize}

The size of these layers is different for each classifier. This is performed so that each layer captures different features which can then be combined to predict the final result. The results of the three classifiers are concatenated to get a vector of length $6$ and passed as input to the final fully connected layer with 2 nodes. The output of this final FC layer is the result of the network.

\subsection{Training and Implementation Details}
The base network and the classifier network are trained together from scratch. The weights in each layer are initialized randomly. The loss function used to train the network is Cross Entropy Loss. The error is backpropagated across all the classifier branches with equal weight and then combined at the base architecture by taking an equally weighted sum. The batch size and initial learning rate while training the model is set to 32 and $1e^{-5}$, respectively. The Adam optimizer \cite{kingma2014adam} is used for training with a weight decay of $0.01$.

\begin{table}[]
    \centering
        \caption{Characteristics of the databases used.}
    \label{tab:databases}
    \begin{tabular}{|c|c|c|c|c|}
    \hline
        Database & Real & Spoof & Sensors & Environment\\
        \hline 
        IIITD-CLI & 2163  & 2165 & 2 & Controlled\\
        \hline
        MUIPAD & 1719  & 1713  & 1 & Uncontrolled\\
        \hline
        UnMIPA & 9319 & 9387 & 3 & Uncontrolled\\
        \hline
    \end{tabular}

\end{table}

\section{Database, Evaluation Criteria and Baseline}
\subsection{Databases}
For testing the proposed MVANet under cross-database, cross-sensor, and cross-environment settings, we perform three-fold cross-validation. For every single fold, the network is trained on one of the three databases and tested on the other two databases. The cross-database testing protocol ensures that a network's performance across unknown sensors and environments can be evaluated. The performance of different networks is compared using total error, APCER, and BPCER. The characteristics of each database are defined below: 

\begin{itemize}
    \item \textbf{IIITD-CLI }\cite{yadav2014unraveling}:  The IIITD Contact Lens Iris (CLI) database contains $6,570$ iris images, which includes the real iris, the textured contact lens, and the transparent (soft) contact lens images. There are $101$ subjects, and since for each of them, both eyes are captured, there are $202$ iris classes. Two different scanners/sensors have been used to capture the iris images, (1) Cogent CIS $202$ dual iris sensor and (2) VistaFA2E single iris sensor. This dataset also addresses the possibility of a particular color being more effective in fooling Iris PAD algorithms by using textured contact lenses of a diverse set of colors like blue, grey, green, and hazel. Since this research focuses on detecting textured contact lenses, we have not used soft lens images during both training and testing. 
    \item \textbf{IIITD-WVU MUIPAD }\cite{yadav2018muipad}: In CLI, all the images are captured in an indoor setting, i.e., a controlled environment. Mobile Uncontrolled Iris Presentation Attack Dataset (MUIPAD) is the first publicly available dataset that has images in both indoor as well as outdoor settings. The images are captured during different times of the day and varying weather conditions to achieve diversity in the dataset. It has $3,432$ real and 3D contact lens iris images corresponding to $70$ iris classes. The IriShield MK2120U mobile sensor is used to capture the images. The dataset has subjects with different ethnicities to increase diversity. To verify whether a particular manufacturer makes lenses that can fool the Iris PAD algorithms, the authors have used lenses by manufacturers like Freshlook, Colorblends, Bausch + Lomb, while still keeping in mind the possible effect of color. 
    \item \textbf{IIITD-WVU UnMIPA} \cite{yadav2019detecting}:  The Unconstrained Multi-sensor Iris Presentation Attack (UnMIPA) Database has $162$ iris classes which are captured using mobile sensors to ensure portability of sensors. This dataset also contains images captured in uncontrolled environment, during different times of the day, which results in different levels of illumination in the same dataset. The authors maintained a mix of colors and manufacturers. In addition to that, three different sensors are used to capture the iris images: (1) EMX-30, (2) BK 2121U, and (3) MK 2120U. In total, the dataset has $18,706$ real and contact lens iris images.  
\end{itemize}
\subsection{Evaluation Criteria}
We use three performance metrics as defined by the ISO standards \cite{iso-pad} to measure the performance of the proposed IPAD algorithm:-
\begin{itemize}
    
    \item Attack Presentation Classification Error Rate (APCER): it is defined as the misclassification rate of attack images being classified into real class. 
    \item Bonafide Presentation Classification Error Rate (BPCER): it is defined as the misclassification rate of real images being classified into attack class. 
    \item Average Classification Error Rate (ACER): it is the average of APCER and BPCER.
\end{itemize}

\subsection{Baseline}
As a baseline approach, the final classification layers of DenseNet \cite{DBLP:journals/corr/HuangLW16a}, ResNet18 \cite{DBLP:journals/corr/HeZRS15}, and VGG16 \cite{DBLP:journals/corr/SimonyanZ14a} are fine-tuned using the training set of each database. The PyTorch \cite{NEURIPS2019_9015} implementations of these networks are used for iris PAD. For our particular use case of binary classification, the final classification layers of all the networks are modified to have only two output nodes. The pre-trained models trained on the ImageNet \cite{imagenet_cvpr09} dataset are used. To use them for iris PAD, the feature extraction layers of these pre-trained models are frozen while the final classification layer is fine-tuned. The models are trained with an Adam optimizer with a learning rate of $0.0001$ and a weight decay of $0.00001$. We use cross-entropy loss criteria to evaluate the training process.



\begin{table}[]
\centering
\caption{Iris presentation attack detection performance (\%) using proposed MVANet under unseen database training and testing conditions. The large-scale database, i.e., UnMIPA yields the best detection results. The best results on each testing database are highlighted.}
\label{tab:proposed-results}
\begin{tabular}{|l|l||c|c|c|c|} \hline
\multirow{1}{*}{Train On} & \multirow{1}{*}{Test On} & \multicolumn{1}{l|}{APCER} & \multicolumn{1}{l|}{BPCER} & \multicolumn{1}{l|}{ACER} & \multicolumn{1}{l|}{Accuracy} \\ \hline
\multirow{2}{*}{IIITD-CLI} & MUIPA & 06.30 & 50.70 & 28.50 & 71.47 \\ \cline{2-6}
 & UnMIPA & 08.13 & 41.94 & 25.03 & 75.03 \\ \hline
\multirow{2}{*}{MUIPA} & IIITD-CLI & 18.68 & 10.60 & 14.64 & 85.36 \\ \cline{2-6}
 & UnMIPA & 26.11 & 06.50 & \textbf{16.31} & 83.66 \\ \hline
\multirow{2}{*}{UnMIPA} & IIITD-CLI & 15.44 & 01.38 & \textbf{08.41} & 91.58 \\ \cline{2-6}
 & MUIPA & 03.33 & 08.91 & \textbf{06.12} & 93.88 \\ \hline
\end{tabular}
\end{table}

\begin{figure}[t]
\begin{center}
   \includegraphics[width=1\linewidth]{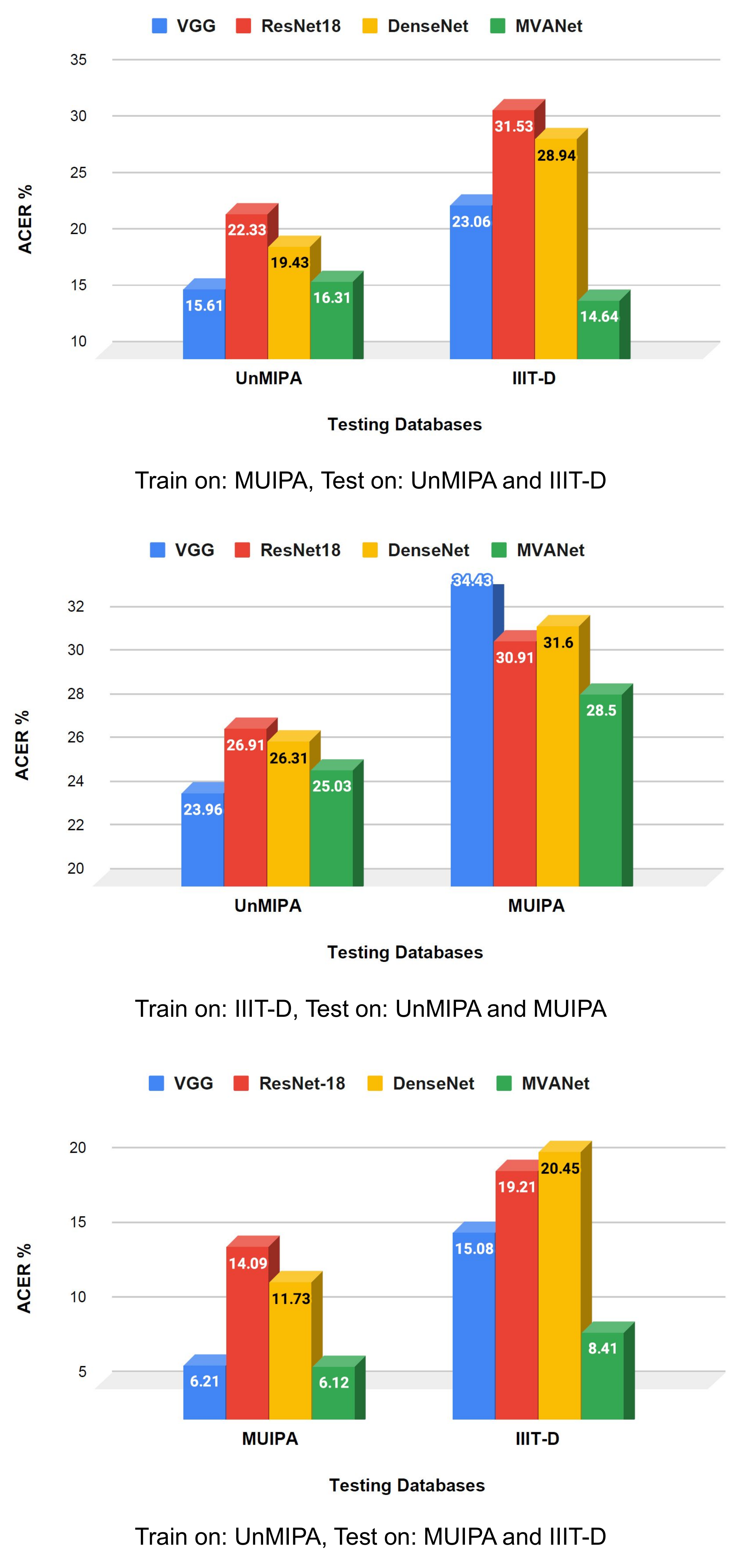}
\end{center}
\caption{Comparing the performance of the proposed iris presentation attack detection algorithm (MVANet) with CNN models under cross database settings. The results are compared using ACER.}    
 \label{fig:compare-bar}
\end{figure}

\begin{figure*}[t]
\begin{center}
   \includegraphics[width=1\linewidth]{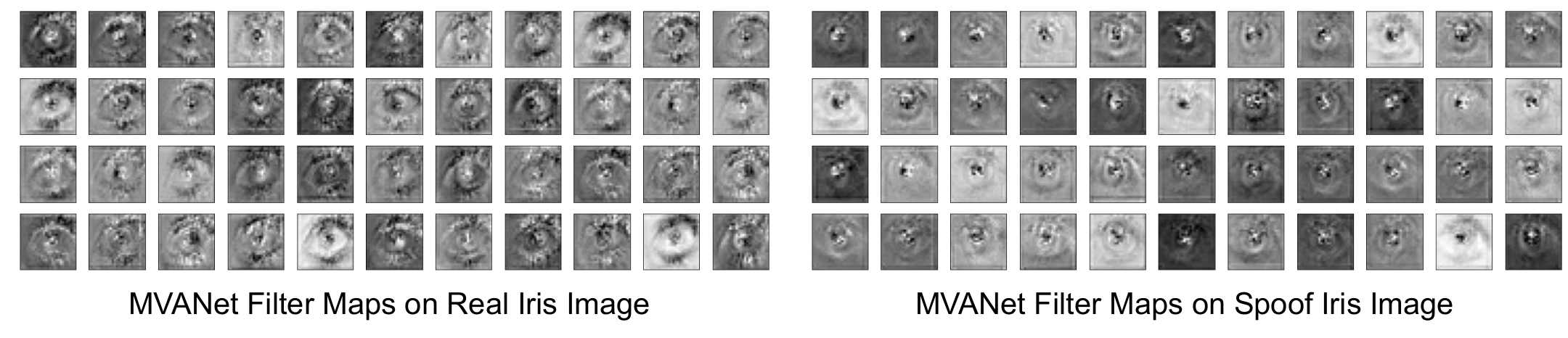}
\end{center}
\caption{Depicting the difference in the CNN filter maps at layer 2 learned over real and spoof iris images.}
\label{fig:filter-maps}
\end{figure*}

\begin{figure*}[t]
\begin{center}
   \includegraphics[width=1\linewidth]{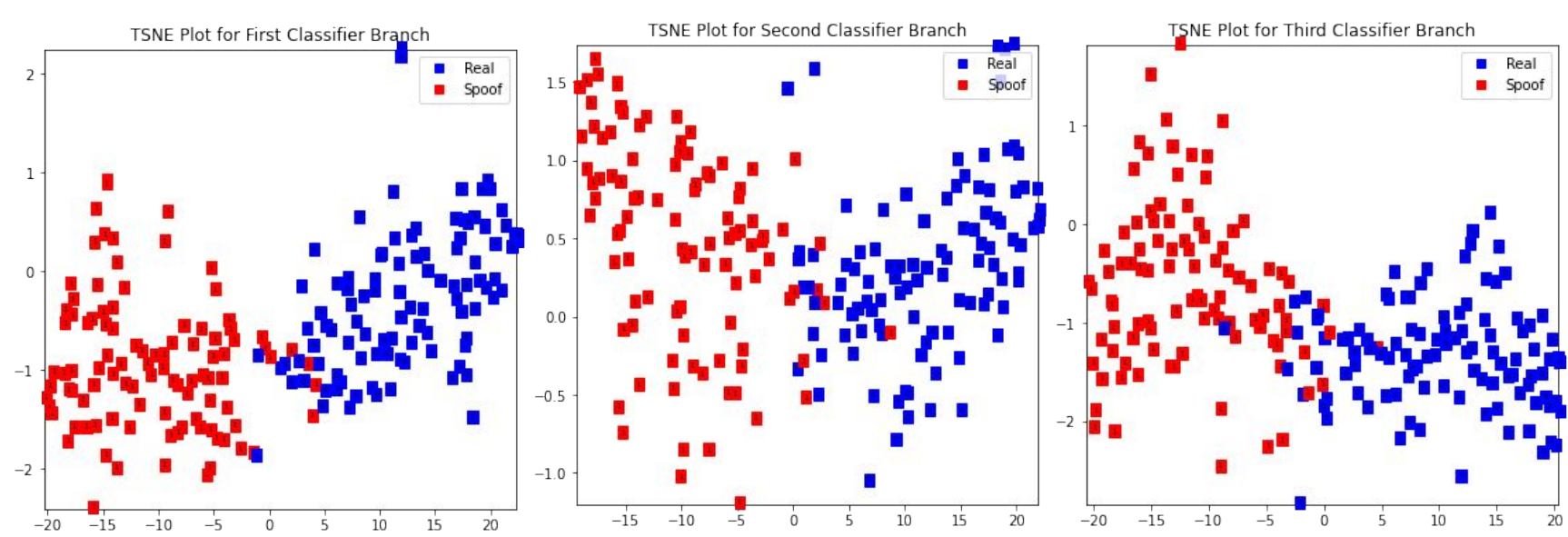}
\end{center}
\caption{TSNE plots for different branches of classifier on 100 iris images, depicting the differences between each branch.}
\label{fig:tsne_plots}
\end{figure*}

\section{Experimental Results}

In this section, the experimental results corresponding to the three-fold training-testing splits are reported using the proposed MVANet and baseline models. When the networks are trained on MUIPAD and tested on UnMIPA and IIIT-CLI, it is observed that the proposed network performs significantly better than VGG16, ResNet and DenseNet, which are fine-tuned for the task of iris PAD. Fig. \ref{fig:compare-bar}(a) helps in visualizing the difference between the performance of the proposed network and baseline approaches. The proposed network yields a significantly lower average ACER on the testing databases as shown in Table \ref{tab:avg_scores}. The challenge of overcoming cross-sensor and cross-environment testing is evident in this case. We observe that all the baseline approaches perform well on only one of the databases. The performance on the other database is significantly inferior. This is where the proposed network performs uniformly across the databases and provides higher accuracy than the other networks. This result is more impressive when we consider the fact that MUIPA consists of samples from only one sensor. At the same time, the other two databases comprise samples from multiple sensors. This result showcases the ability of the network to work well across different sensors.


A similar trend in performance is observed when the networks are trained on the UnMIPA database and tested on MUIPA and IIIT-CLI. In this case, however, the performance of all the networks is decent. Fig. \ref{fig:compare-bar}(b) shows that the proposed network performs significantly better than all the baseline networks with cross-sensor and cross-environment scenarios. The reason for all the networks performing well when trained on UnMIPA can be explained by the training size, the high variation in the environment, and the use of multiple sensors. 
\begin{table}[]
    \centering
        \caption{Comparison of the proposed MVANet with three CNN architectures in terms of average ACER scores when a particular database is used for training and remaining are used for testing. For example, when MVANet is trained using UnMIPA, it yields an ACER value of $8.41$\% on IIITD-CLI and $6.12$\% on MUIPA. Therefore, the average of the two is $\mathbf{7.26}$\%. Bold represents the best results.}
    \label{tab:avg_scores}
    \begin{tabular}{|c|c|c|c|c|}
    \hline
        Training Database & DenseNet & ResNet & VGG16 & MVANet\\
        \hline 
        IIITD-CLI & 28.95 & 28.91 & 29.19 & \textbf{26.76}\\
        \hline
        MUIPAD & 24.19  & 26.93  & 19.34 & \textbf{15.47}\\
        \hline
        UnMIPA & 16.09  & 16.65 & 10.65 & \textbf{07.26}\\
        \hline
    \end{tabular}

\end{table}

\begin{figure}[]
\begin{center}
   \includegraphics[width=1\linewidth]{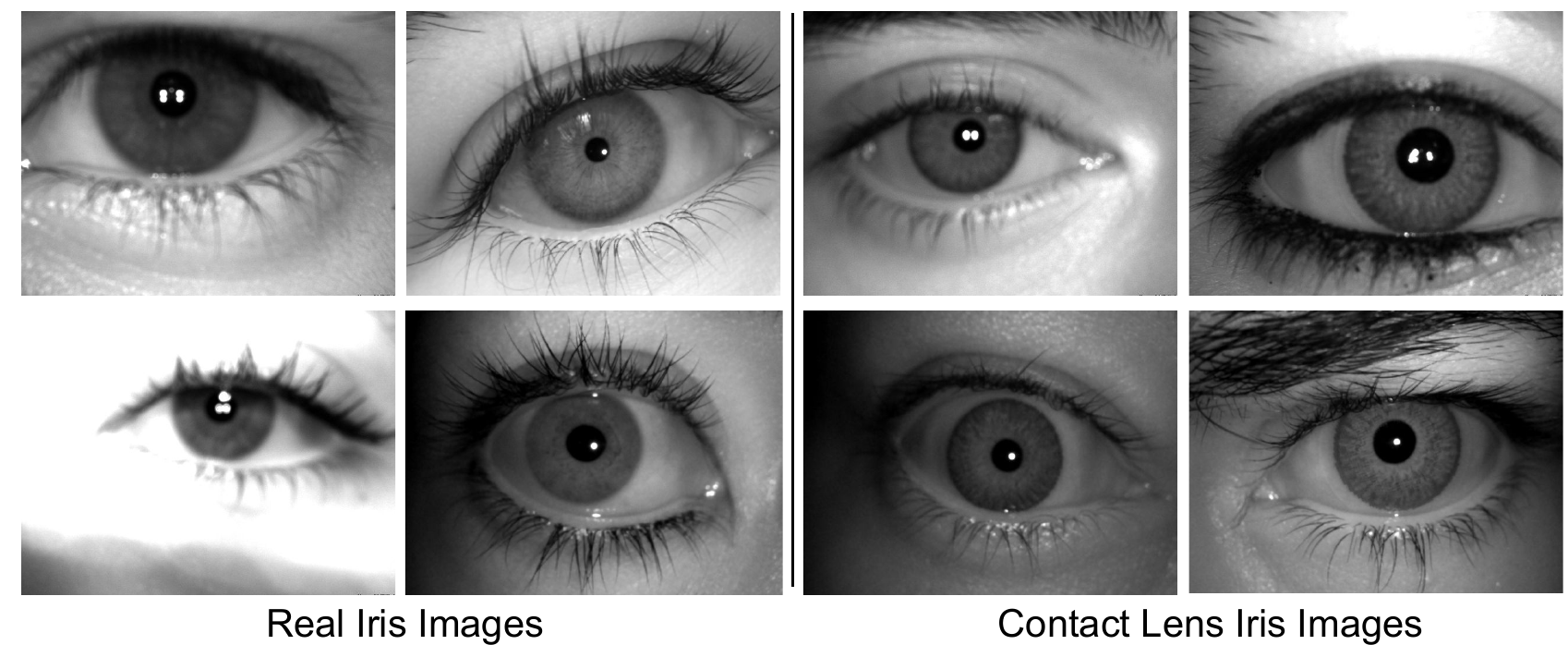}
\end{center}
\caption{Real and contact lens iris images misclassified by baseline networks but correctly classified by MVANet. Figure depicts that MVANet is effective in handling the variations present in the iris images corresponding to different databases and sensors.}
\label{fig:misclassified}
\end{figure}

\begin{figure}[]
\begin{center}
   \includegraphics[width=1\linewidth]{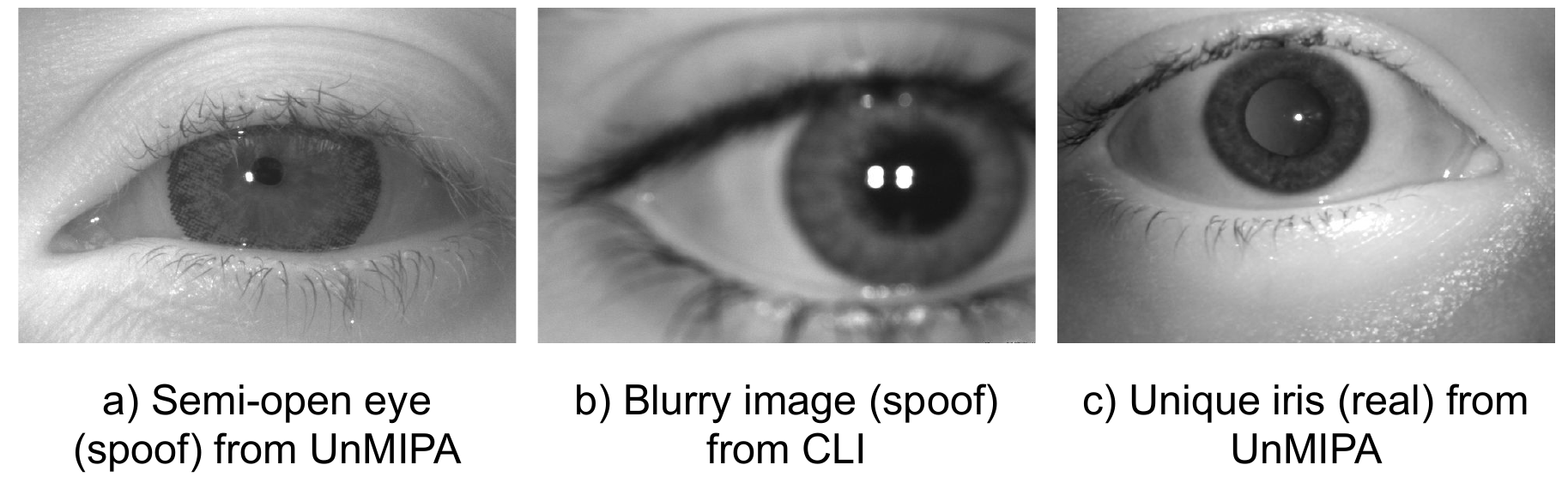}
\end{center}
\caption{Samples of real and contact lens iris images misclassified by MVANet.}
\label{fig:incorrect}
\end{figure}

Training the networks on the IIITD-CLI database yields a significant result for our problem. The samples in this database are all collected in a controlled environment, i.e., all the images are captured indoors. This case highlights the difficulties faced in cross-environment testing where a network trained in a controlled environment fails to perform well on samples from an uncontrolled environment. The proposed network, however, still outperforms all the baseline approaches, as shown in Fig. \ref{fig:compare-bar}(c). The consistent and better performance of the proposed network, paired with its much lesser training time, makes it an ideal network for the task of iris-PAD for multi-sensor and multi-environment tasks.

\subsection{Analysis}



In this section, we analyze the cases where the proposed MVANet architecture performs better in comparison to other deep learning based architectures. Fig. \ref{fig:misclassified} shows examples of real and presentation attack cases which are incorrectly classified by DenseNet \cite{DBLP:journals/corr/HuangLW16a} but correctly classified by the proposed architecture. These cases show the variations due to illumination, lens colors, and eye colors, and highlight the importance of learning different kinds of features specific to the task of iris PAD. In cross-sensor and cross-environment testing, the APCER for DenseNet, is 7.47\%, and the BPCER is 16.01\%. Using the same protocol, the proposed network yields APCER of 3.33\% and BPCER of 8.91\%. This significant drop in both the error rates can be attributed to the proposed network's ability to learn different kinds of features. 

Fig. \ref{fig:filter-maps} shows the filter maps of real and spoofed iris images obtained from MVANet (at layer 2). 
The difference between these filters is clear from the maps. For real iris images, the filters capture more edge information. For spoof iris images, maps are concentrated mostly around the center shown by the black and white circles at the center of the maps. Fig. \ref{fig:tsne_plots} highlights the advantage of using three different classifiers after feature extraction. As observed in the plots, different branches can capture different information in the images, which helps the proposed network to perform significantly better in cross-database settings.


Fig. \ref{fig:incorrect} shows some of the images misclassified by the proposed MVANet. As shown in  Fig. \ref{fig:incorrect}(a), in some cases, since the eye is not properly open, the proposed algorithm makes an incorrect decision. In such cases, we observe that the entire iris is not exposed to the camera, thus, making it difficult to classify the image as real or spoof. As shown in Fig. \ref{fig:incorrect}(b), another covariate for the proposed network is handling out-of-focus samples. Fig. \ref{fig:incorrect}(c) captures an unusual case leading to misclassification, where the real iris of the person is very different from the common eye. It is because of such a drastic difference that the proposed network classifies this image as a spoof image. Handling such cases would require a larger variety of iris types to train the network.

\noindent \textbf{Computational Complexity:}
All the experiments are performed using Google's Colab platform. The system is powered by Intel's Xeon processors and Nvidia's K80 GPUs. We train all the networks on the GPUs for faster training. For comparing the training time, we train the networks on the UnMIPA database, which contains $18,706$ images. It is essential to state that MVANet is trained from scratch while the baseline networks are just fine-tuned. The proposed algorithm required an average of $6$ minutes and $47$ seconds per epoch. VGG16 and DenseNet required $7$ minutes and $41$ seconds per epoch, and $7$ minutes and $52$ seconds per epoch, respectively. The proposed network takes lesser time to train entirely than the baseline networks' fine-tuning. This lower training time, paired with better cross-database results, makes the proposed network an ideal choice for real-time iris PAD implementation.

\subsection{Intra-Database Experimental Results}

As mentioned in the literature review section, most of the existing work has followed the intra-database training-testing experimental protocol. Therefore, we have also performed the experiments using the intra-database protocol defined by Yadav et al. \cite{yadav2014unraveling} on IIITD-CLI. The experiments are performed on individual sensors images. The results of the proposed MVANet along with existing hand-crafted and CNN based algorithms are summarized in Table \ref{intra-iiitd}. The proposed algorithm outperforms existing algorithms based on hand-crafted features along with three CNN models used for IPAD. The hand-crafted features used for comparison are: mLBP \cite{yadav2014unraveling}, Textural features \cite{wei2008counterfeit}, LBP \cite{ojala1996comparative} with SVM, weighted LBP \cite{zhang2010contact}, and fusion of LBP and PHOG \cite{bosch2007representing}. The results demonstrate that the the proposed algorithm yields significantly high detection accuracy compared to existing algorithms. For instance, MVANet yields $4.50$\% and $0.29$\% higher accuracy than the second-best algorithm, i.e., VGG for cogent and vista sensor images detection, respectively. In comparison to hand-crafted features, the performance of the proposed MVANet is at-least $14.03$\% (MVANet vs. mLBP \cite{yadav2014unraveling}) and $8.05$\% (MVANet vs. textural features \cite{wei2008counterfeit}) better on cogent and vista images, respectively. Overall, it is observed that iris images captured using cogent senors are challenging to detect compared to the vista sensor.

\begin{table}[t]
\centering
\caption{Iris presentation attack detection accuracy (\%) of the proposed MVANet, CNNs, and existing algorithms in intra-database training-testing. Bold represents the best results.}
\label{intra-iiitd}
\begin{tabular}{|l|c|c|}\hline
Algorithm & {Cogent} & {Vista} \\ \hline
Textural Features \cite{wei2008counterfeit} & 55.53 & 87.06 \\ \hline
Weighted LBP & 65.40 & 66.91 \\ \hline
LBP + SVM & 77.46 & 76.01 \\ \hline
LBP + PHOG + SVM & 75.80 & 74.45 \\ \hline
mLBP \cite{yadav2014unraveling} & 80.87 & 83.91 \\ \hline
ResNet18 & 85.15 & 80.97 \\ \hline
DenseNet & 84.32 & 91.83 \\ \hline
VGG & 90.40 & 94.82 \\ \hline
Proposed MVANet & \textbf{94.90} & \textbf{95.11} \\ \hline
\end{tabular}
\end{table}

\section{Conclusion}

Textured contact lenses have been known to affect the performance of iris recognition systems. Existing attack detection algorithms have been demonstrated to be effective in seen environments; however, in new/unseen environments for instance unseen sensors or attacks, the error rates are significantly higher which makes the deployment of these systems unreliable. In this paper, we present a deep learning based architecture termed as MVANet which is agnostic to database, acquisition sensors, and imaging environments. Experiments conducted on the IIIT-WVU UnMIPA, IIITD-WVU MUIPA, and IIITD CLI databases show that the proposed network is not only effective in intra-database and cross-database settings but it is also computationally efficient. In the future, the aim is to further improve the detection performance even with limited number of training images. We also plan to extend the proposed algorithm to work under adverse conditions including presence of medical disorders or under the influence of alcohol and drugs \cite{ssr}.

\section*{Acknowledgment}

A. Agarwal is partly supported by the Visvesvaraya PhD Fellowship. R. Singh and M. Vatsa are partially supported through a research grant from MeitY, India. M. Vatsa is also partially supported through the Swarnajayanti Fellowship by the  Government of India.

\bibliographystyle{ieee_fullname}
\bibliography{1453}

\end{document}